\documentclass[nohyperref]{article}

\usepackage{microtype}
\usepackage{graphicx}
\usepackage{subfigure}
\usepackage{booktabs} 

\usepackage{hyperref}

\usepackage[accepted]{icml2022}

\usepackage{amsmath}
\usepackage{amssymb}
\usepackage{mathtools}
\usepackage{amsthm}

\usepackage{makecell}   

\usepackage[capitalize,noabbrev]{cleveref}

\theoremstyle{plain}

\theoremstyle{definition}

\theoremstyle{remark}

\usepackage[textsize=tiny]{todonotes}

\icmltitlerunning{Pre-training Transformers for Molecular Property Prediction Using Reaction Prediction}

\begin{document}

\twocolumn[
\icmltitle{Pre-training Transformers for Molecular Property Prediction Using Reaction Prediction}



\icmlsetsymbol{equal}{*}

\begin{icmlauthorlist}
\icmlauthor{Johan Broberg}{comp}
\icmlauthor{Maria Bånkestad}{comp,yyy}
\icmlauthor{Erik Ylipää}{comp}
\end{icmlauthorlist}

\icmlaffiliation{comp}{Research Institute of Sweden (RISE), Isafjordsgatan 22, 164 40 Kista, Sweden}
\icmlaffiliation{yyy}{Department of Information Technology, Uppsala University, Uppsala, Sweden}

\icmlcorrespondingauthor{Johan Broberg}{johan.broberg@ri.se}

\icmlkeywords{Machine Learning, ICML, Transformer, Molecular Property Prediction, Transfer Learning, Pre-training}

\vskip 0.3in
]



\printAffiliationsAndNotice{}  

\begin{abstract}
Molecular property prediction is essential in chemistry, especially for drug discovery applications. However, available molecular property data is often limited, encouraging the transfer of information from related data. Transfer learning has had a tremendous impact in fields like Computer Vision and Natural Language Processing signaling for its potential in molecular property prediction. We present a pre-training procedure for molecular representation learning using reaction data and use it to pre-train a SMILES Transformer. We fine-tune and evaluate the pre-trained model on 12 molecular property prediction tasks from MoleculeNet within physical chemistry, biophysics, and physiology and show a statistically significant positive effect on $5$ of the $12$ tasks compared to a non-pre-trained baseline model.
\end{abstract}

\section{Introduction}
\label{introduction}

Molecular property prediction has long been used to quickly screen new molecule leads in drug development. The accuracy of these methods is crucial, since false negatives incur high costs when a lead is taken to an experimental phase. Lately, machine learning has become one of the standard tools for molecular property prediction. However, the main challenge is the limited amount of available data to train models on. One solution to this problem is to curate larger datasets using domain expertise. This can be a costly and time consuming approach but has the advantage that larger parts of the relevant molecular domain can be covered~\cite{photoswitch}. Another approach to solve the data scarcity problem is that of \emph{transfer learning}~\citep{transfer_learning}, where knowledge in one task is used to improve performance in another. This is an active area of research in the chemistry domain, mainly using Graph Neural Networks (GNNs)~\citep{Kipf} and Transformer models~\citep{AttentionIsAllYouNeed}. While transfer learning has proven to be very successful in domains such as Natural Language Processing~\citep{ulmfit, BERT, RoBERTa} and Computer Vision~\citep{CV_transfer_learning}, the same clear success is yet to happen in chemistry.




This work is a continuation of the master's thesis \cite{Broberg2022-lh} and explores a pre-training strategy for molecular representation learning based on chemical reaction prediction. We use it to pre-train a transformer encoder and compare its performance to a randomly initialized one on a wide range of molecular property prediction tasks. We show statistically significant improvements on $5$ of the $12$ datasets using a significance level $\alpha = 0.05$ with Bonferroni correction~\citep{bonferroni1935calcolo, bonferroni1936teoria, MultipleTestingCorrection}.

\begin{figure*}[t]
    \centering
    \includegraphics[width=0.55\linewidth]{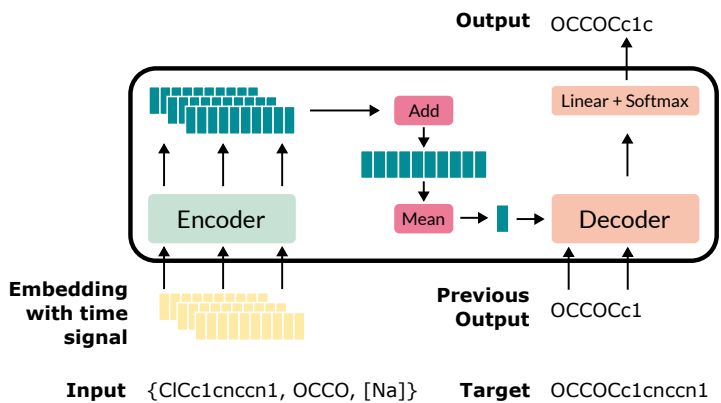}
    \caption{Illustration of our pre-training architecture. Note that each molecule fragment is encoded independently of the others.}
    \label{fig:my_transfer_learning}
\end{figure*}

\section{Related Work}
Most recent work on pre-training deep models for molecular property prediction uses either GNNs or Transformers.

With GNNs, it is common to use multiple learning objectives that aims to improve representation on different levels (node/edge/graph)~\citep{strategies_for_pretraining_GNNs, grover,pretraining_3D}. Node and edge level pre-training tasks generally aim to capture graph structural regularities of molecules. Examples of such tasks are prediction of masked node or edge attributes or using node embeddings to predict information about the neighborhood structure. Graph level tasks may also be based on graph structural information but there are also approaches that more explicitly utilize information from the chemistry domain. For example, 3D molecular structure data~\citep{3D_infomax, Geometry_enhanced, pretraining_3D} and graph motifs with their connections to functional groups~\citep{grover, motif_based} have been used for graph level pre-training.

For Transformers applied to molecular property prediction using SMILES~\citep{SMILES}, a common pre-training approach is to randomly mask parts of the input string. ChemBERTa~\citep{ChemBERTa}, MolBERT~\citep{molbert}, and SMILES-BERT~\citep{SMILES_BERT} are only some of the works that explore this method. Though proven beneficial, this approach has not led to the huge improvements seen in NLP by, e.g., the BERT model~\citep{BERT} using the very related masked-language-modeling (MLM) pre-training task. \citet{ChemBERTa} show a diminishing performance gain on three tasks (BBBP, ClinTox (CT\_TOX), and Tox21 (SR-p53)) using a masked token prediction pre-training strategy where an increase in dataset size from $10^6$ to $10^7$ only lead to a $\approx 3\%$ mean ROC-AUC increase and a $\approx 2\%$ mean PRC-AUC increase. This indicates that pre-training by recovering masked tokens alone might not scale well enough to train powerful property prediction models.

Another recent line of research tries to adapt the Transformer architecture to take molecular graphs as input~\citep{MAT, GRAT, Graphormer}. Such modifications allow Transformers explicit access to the graph structure, which otherwise must be learned implicitly from string representations. Furthermore, it can also allow information such as node features and edge features to be included in the model input. These models have achieved remarkable results and have in some cases been pre-trained using structure-based tasks~\citep{MAT, GRAT}, but the main contribution to their predictive power seem to stem from their architecture rather than their choice of pre-training task.

A relatively unexplored source of information for pre-training in the chemistry domain is reaction data. To our knowledge, the only published work using such data for pre-training is by \citet{chemical_reaction_aware_pretraining}. They base their approach on GNNs and a contrastive learning task that teaches the model to encode molecules such that two aggregated sets of molecular encoding vectors lie near each other in the encoding space if they make up the left- and right-hand side of a chemical reaction respectively, and far away if they do not. 

\begin{figure}[b]
    \centering
    \includegraphics[width=\linewidth]{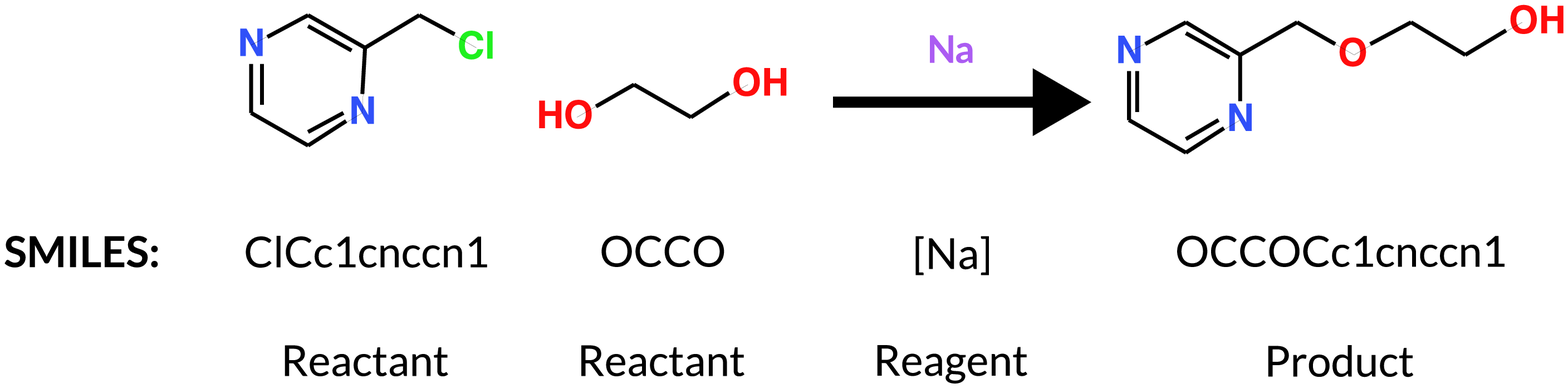}
    \caption{Example reaction with molecules represented as their structural formulas and as SMILES.}
    \label{fig:reaction}
\end{figure}

Our approach differs from that of \citet{chemical_reaction_aware_pretraining} in that we model the pre-training step as a generative reaction prediction task instead of a contrastive learning task. Furthermore, we use a transformer architecture and SMILES representations of molecules while they use a GNN architecture and graph representations.

\section{Background}

\subsection{Chemical Reactions}
A chemical reaction is a transformation of one set of molecules into another. The molecules present before the reaction are called \textit{reactants}, while the molecules created through the reaction are called \textit{products}. Molecules might be part of the reaction but not contribute themselves with any atoms to the product molecules created. These are called \textit{reagents}. In reaction prediction one tries to predict the product molecules of a reaction given the reactants and reagents. Normally, a reaction produces multiple product molecules. In our work, we have limited the scope by using data with only a single product.

\subsection{SMILES}
SMILES, Simplified-Molecular-Input-Line-Entry-System is a linearization of the molecular graph. A molecule has many possible SMILES depending on where the linearization starts and what branches to take. There are certain rules for producing \textit{canonical} SMILES of molecules, where the same molecule will always encode to the same SMILES. A commonly used strategy when using SMILES as inputs to neural networks is to \textit{randomize} the SMILES, by essentially starting the linearization at a random place and take random branches when traversing the molecular graph. This has shown to act as a powerful data augmentation method when working with SMILES representations of molecules~\citep{randomized_smiles,arus2019randomized}.

\subsection{Transformer}

The Transformer is an architecture operating on mathematical sets and was introduced by \citet{AttentionIsAllYouNeed} in the context of neural machine translation. It has been widely used for sequential data such as natural language. A positional encoding is then added to the input data to provide the sequential structure to the model. The original (full) Transformer consists of an encoder and a decoder and is typically used for translation tasks. On its own, the Transformer encoder can be used for sequence classification/regression/representation tasks~\citep{BERT}. The key component of the architecture is the multi-head attention mechanism which enables the model to attend to all elements of its input at once. For a detailed description of the full Transformer model and for Transformer encoder models such as BERT we refer to \citep{AttentionIsAllYouNeed} and \citep{BERT} respectively.

\section{Method}

Our approach is based on the Molecular Transformer by \citet{MolecularTransformer}, in which reaction prediction is modeled as a sequence translation problem for which a full Transformer model is used. Like \citet{MolecularTransformer}, we represent molecules as SMILES. By using canonicalized SMILES for our product molecules we obtain a fixed target sequence which makes the generative process easier compared to generating graphs, where an order needs to be induced on the edge set~\citep{ordermatters}.

\subsection{Pre-Training Architecture}
In the Molecular Transformer, the encoder is applied to SMILES strings containing all reactants and reagents of the corresponding reaction. In our pre-training phase, the Transformer encoder is applied to each reactant and reagent independently. That is, the encoder can only attend to the tokens in the same SMILES fragment, not across fragments. For the set of reactants and reagents $R = \{r_1, r_2, ...\}$ in a given reaction, each SMILES fragment $r_i$ produces an encoding $\mathbf{h}_i \in \mathbb{R}^{L \times d}$ where $L$ is the maximum sequence length and $d$ is the token embedding dimension. The set of such encodings $H = \{\mathbf{h}_1, \mathbf{h}_2, ...\}$ are then aggregated into a single vector $h_{R} \in \mathbb{R}^{d}$ representing the entire reaction by first applying element-wise addition across the encodings and then averaging across the sequence axis according to

\begin{equation}
    h_R = \text{Aggregate}(H) = \text{Mean}(\text{Sum}(H)).
\end{equation}

The aggregated reaction vector $h_R$ is then passed to the decoders ``encoder-decoder attention'' layers as a memory key to all tokens in the product (target) SMILES. 

We average across the sequence axis because, in the downstream fine-tuning tasks, we will use encoded molecules together with a Multi-Layered Perceptron (MLP). This mean that we will also need to aggregate the encoded molecules into fixed sized vectors. We choose to include this aggregation in the pre-training step, so that the single vector representation will be forced to contain all information needed for the decoding. 

\subsection{Fine-Tuning Architecture}
In the fine-tuning phase, we only use the encoder component from the pre-training phase. Encodings of molecules are aggregated across the sequence axis using the mean. A 2-layered MLP with ReLU activations is used to map the aggregated molecular encodings to the target values. All parameters (encoder and MLP) are tuned in this phase.

\begin{figure}[t]
    \centering
    \includegraphics[width=0.45\linewidth]{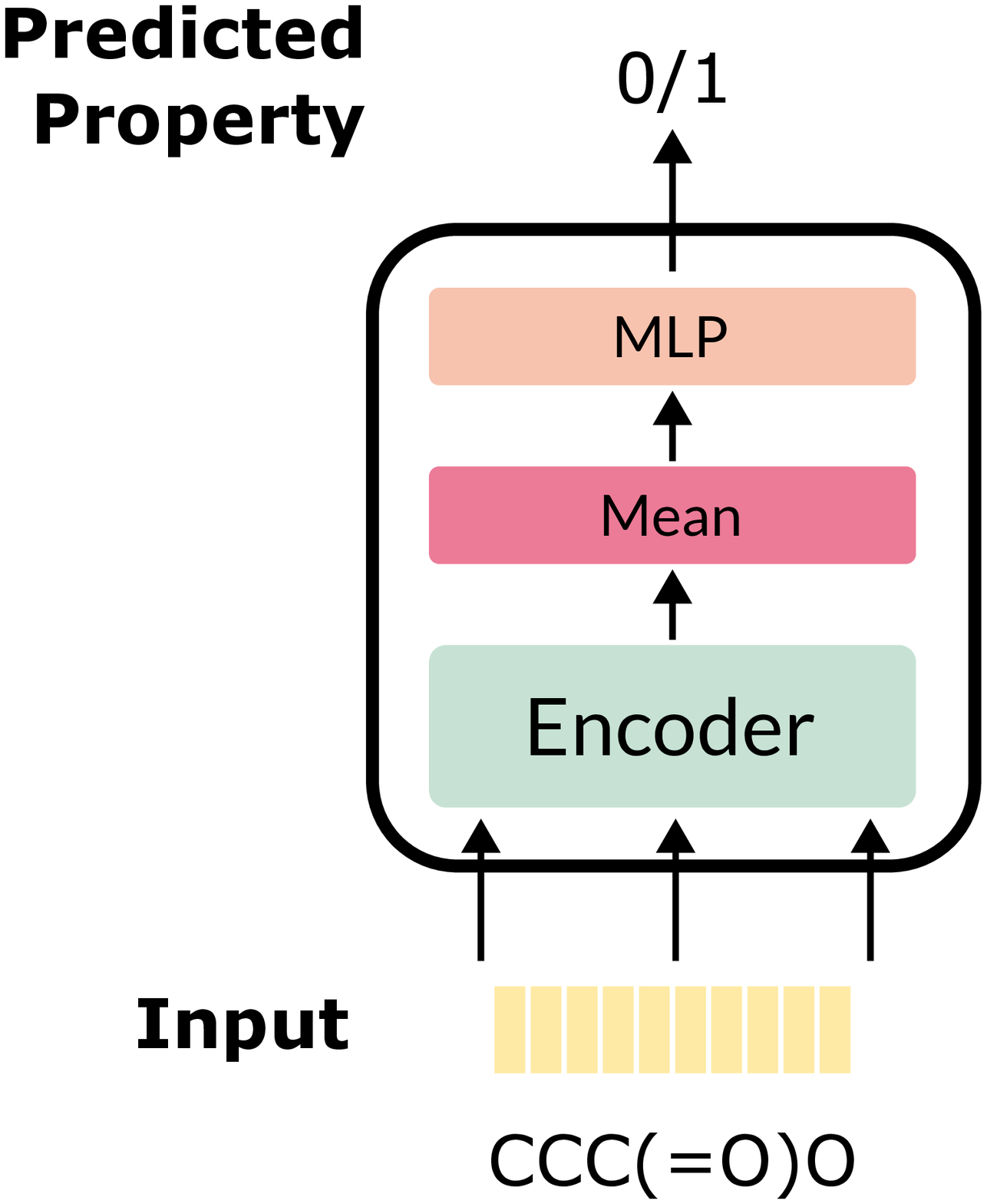}
    \caption{Illustration of the fine-tuning architecture for a binary classification task.}
    \label{fig:my_fine_tuning}
\end{figure}

\subsection{Evaluation}

\begin{table*}[t]
    \caption{Performance of the model with and without pre-training showing the mean and standard deviation over a 10-fold cross-validation.}
    \label{tab:results}
    \vskip 0.15in
    \begin{center}
    \begin{small}
    \begin{sc}
        \begin{tabular}{lrrrcc}
             \toprule
             Data set & Metric & \makecell{Model\\Without\\pre-training} & \makecell{Model\\With\\pre-training} & \makecell{Wilcoxon\\Signed-Rank\\Test} & \makecell{Rank-Biserial\\Correlation}\\ \midrule
             ESOL & RMSE $\downarrow$ & $0.656 \pm 0.082$ & $\mathbf{0.428} \pm 0.077$ & $\mathbf{0.001}$ & $1.000$\\
             FreeSolv & RMSE $\downarrow$ & $2.057 \pm 0.477$ & $\mathbf{1.484} \pm 0.413$ & $\mathbf{0.002}$ & $0.982$\\
             Lipophilicity & RMSE $\downarrow$ & $1.012 \pm 0.038$ & $\mathbf{0.700} \pm 0.035$ & $\mathbf{0.001}$ & $1.000$\\ \midrule
             PCBA & PRC-AUC $\uparrow$ & $0.178 \pm 0.010$ & $0.175 \pm 0.006$ & $0.862$ & $0.309$\\
             MUV & PRC-AUC $\uparrow$ & $0.023 \pm 0.026$ & $0.025 \pm 0.017$ & $0.539$ & $0.491$\\
             \midrule
             HIV & ROC-AUC $\uparrow$ & $0.704 \pm 0.050$ & $\mathbf{0.757} \pm 0.052$ & $\mathbf{0.002}$ & $0.982$\\
             BACE & ROC-AUC $\uparrow$ & $0.726 \pm 0.088$ & $\mathbf{0.817} \pm 0.106$ & $\mathbf{0.001}$ & $1.000$\\
             BBBP & ROC-AUC $\uparrow$ & $0.902 \pm 0.104$ & $0.921 \pm 0.101$ & $0.019$ & $0.873$\\
             Tox21 & ROC-AUC $\uparrow$ & $0.799 \pm 0.012$ & $0.792 \pm 0.013$ & $0.981$ & $0.145$\\
             ToxCast & ROC-AUC $\uparrow$ & $0.693 \pm 0.021$ & $0.701 \pm 0.013$ & $0.348$ & $0.582$\\
             SIDER & ROC-AUC $\uparrow$ & $0.597 \pm 0.016$ & $0.578 \pm 0.039$ & $0.920$ & $0.255$\\
             ClinTox & ROC-AUC $\uparrow$ & $0.956 \pm 0.0044$ & $0.959 \pm 0.038$ & $0.652$ & $0.436$\\
             \bottomrule
        \end{tabular}
    \end{sc}
    \end{small}
    \end{center}
    \vskip -0.1in
\end{table*}

We fine-tune separately on 12 datasets in MoleculeNet~\citep{MoleculeNet} using 10-folded cross-validation. Each fold ($10\%$) is used once for evaluation of hyperparameter tuning, once for validation, once for testing and otherwise for training. For regression tasks and multi-label classification tasks we use random splits while for single-label classification tasks we use stratified splits. On multi-label prediction benchmarks (PCBA, MUV, TOX21, ToxCast, SIDER, ClinTox) we report the average performance across all tasks as suggested by \citet{MoleculeNet}.

To evaluate the effect of our transfer learning approach, we compare the pre-trained model to a randomly initialized one which we train directly on each molecular property dataset. Evaluation is based on the best performing checkpoints with respect to the validation set, for each model run. Our null hypothesis is that reaction prediction pre-training has no effect on molecular property prediction across all chemical space. The null hypothesis is then rejected  with 95\% level of confidence.

We pair the performances on each test fold and use the Wilcoxon signed-rank test~\citep{wilcoxon} to determine statistically significant differences between our pre-trained model and the randomly initialized one on each of the 12 datasets. The Wilcoxon signed-rank test is a non-parametric version of the Student's t-test, which does not assume normally distributed data. Since we evaluate using multiple tests, we also make a Bonferroni correction to counteract the \textit{multiple comparison problem}~\citep{MultipleTestingCorrection}. Practically this means that, to evaluate our null hypothesis with a 95\% level of confidence (significance level $\alpha = 0.05$) we use a significance level of $\alpha_1 = ... = \alpha_{m} = \alpha / m$ when we test for an effect on each individual dataset. Here $m = 12$ and denote the number of datasets we test on. This means that we for each dataset we use a significance level of  $\alpha_1 = ... = \alpha_{12} = 0.05/12 = 0.00417$.

\section{Experiment}

\subsection{Data and Pre-Processing}
The dataset used in the pre-training phase is from the USPTO database~\citep{Lowe2012} and consists of $902,581$ samples used for training and $50,131$ samples used for validation, based on pre-processing and data splits provided by \citet{FoundInTranslation}. 

For the reactants and reagents we use randomized SMILES while the product molecules, our targets, were kept in canonicalized form. Due to the memory complexity of the Transformer we truncate SMILES in the training data to a maximum sequence length of 157. Of the reactions in the pre-training data, $99.9\%$ contain reactants, reagents and products whose SMILES are all shorter than 157, so this truncation threshold only affects $0.1\%$ of the reactions. 

\subsection{Experimental Setup}
For our pre-training model we used a four-layered encoder and decoder, with eight attention heads and a layer width of 256. We pre-trained our model for 150 epochs, using cross-entropy loss and AdamW optimizer with a batch size of $4096$ and a cosine cyclic learning rate scheduler with base learning rate of $10^{-5}$ and maximum learning rate of $5 \cdot 10^{-4}$. 

In the fine-tuning phase we tuned the learning rate for all models on each fold in the cross validation. We did this in order to fairly compare the performance. Each tuning was based on 20 runs with learning rates sampled geometrically in the interval $\left[10^{-6}, 10^{-3}\right]$. During learning rate tuning we trained each run for $50$ epochs for all datasets except PCBA and MUV which were only tuned for one respectively ten epochs due to the large number of samples in these datasets. The batch size was set to $64$. For the number of epochs, we used early stopping with $40$ update steps of patience. Throughout this work we tokenized SMILES by converting each character to their corresponding ASCII-value.

\subsection{Results}

The results from our experiment are shown in table \ref{tab:results}. For each dataset we present the mean and standard deviation over the 10-folded cross-validation along with the $p$-value of the Wilcoxon signed-rank test and the rank-biserial correlation. On 5 of the 12 downstream property prediction tasks our pre-training strategy show a statistically significant positive effect given the Bonferroni-corrected significance level $\alpha = 0.00417$. We therefore reject our null hypothesis and conclude that our pre-training approach using reaction prediction has an effect on molecule property prediction

\section{Limitations}
This pre-training strategy has two limitations that we would like to point out. We use reactions that only have one major product molecule, but most reactions contain more than one product. This is an advantage of the approach proposed by \citet{chemical_reaction_aware_pretraining}. Another limitation is that we base our statistical analysis on cross-validation of the downstream tasks. A more robust statistical analysis would have been based on several different runs of pre-training.

\section{Conclusions}

In this paper we have presented a pre-training strategy for transformer models using reaction prediction. We demonstrate a statistically significant effect on 5 out of 12 datasets from MoleculeNet and conclude that reaction prediction pre-training has an effect on molecular property prediction. Our results show, in line with \citet{chemical_reaction_aware_pretraining}, that chemical reactions can be used to successfully pre-train models for downstream molecular property prediction tasks.

\bibliography{refs}
\bibliographystyle{icml2022}

\end{document}